\documentclass{article}
\usepackage{amssymb}

\usepackage{graphicx} 
\usepackage{caption}
\usepackage{amsmath} 
\usepackage[final]{corl_2025} 
\usepackage{array}
\usepackage{booktabs} 
\usepackage{multirow}
\usepackage{wrapfig}
\usepackage{tabularx}
\usepackage{dblfloatfix}
\usepackage{amssymb,placeins}
\usepackage{wrapfig} 
\usepackage{titlesec}  
\titleformat{\subsection}{\normalfont\large\bfseries}{\thesubsection}{1em}{} 
\titlespacing*{\subsection}{0pt}{3pt}{5pt} 

\newcolumntype{C}[1]{>{\centering\arraybackslash}m{#1}}

\makeatletter
\newcommand{\customsize}{\@setfontsize\customsize{8.5pt}{9pt}} 
\makeatother

\title{OPAL: Visibility-aware LiDAR-to-OpenStreetMap Place Recognition via Adaptive Radial Fusion}

\author{
  Shuhao Kang\textsuperscript{1}\textsuperscript{*}
  Martin Y. Liao\textsuperscript{2}\textsuperscript{*}
  Yan Xia\textsuperscript{3\textdagger}
  Olaf Wysocki\textsuperscript{4}
  Boris Jutzi\textsuperscript{1,5}
  Daniel Cremers\textsuperscript{1} \\[10pt]
  \textsuperscript{1} Technical University of Munich 
  \textsuperscript{2} Wuhan University \\
  \textsuperscript{3} University of Science and Technology of China
  \textsuperscript{4} University of Cambridge
  \textsuperscript{5} Karlsruhe Institute of Technology
}

\begin{document}
\maketitle

\begin{figure}[htbp]
  \centering
  \includegraphics[width=1.0\linewidth]{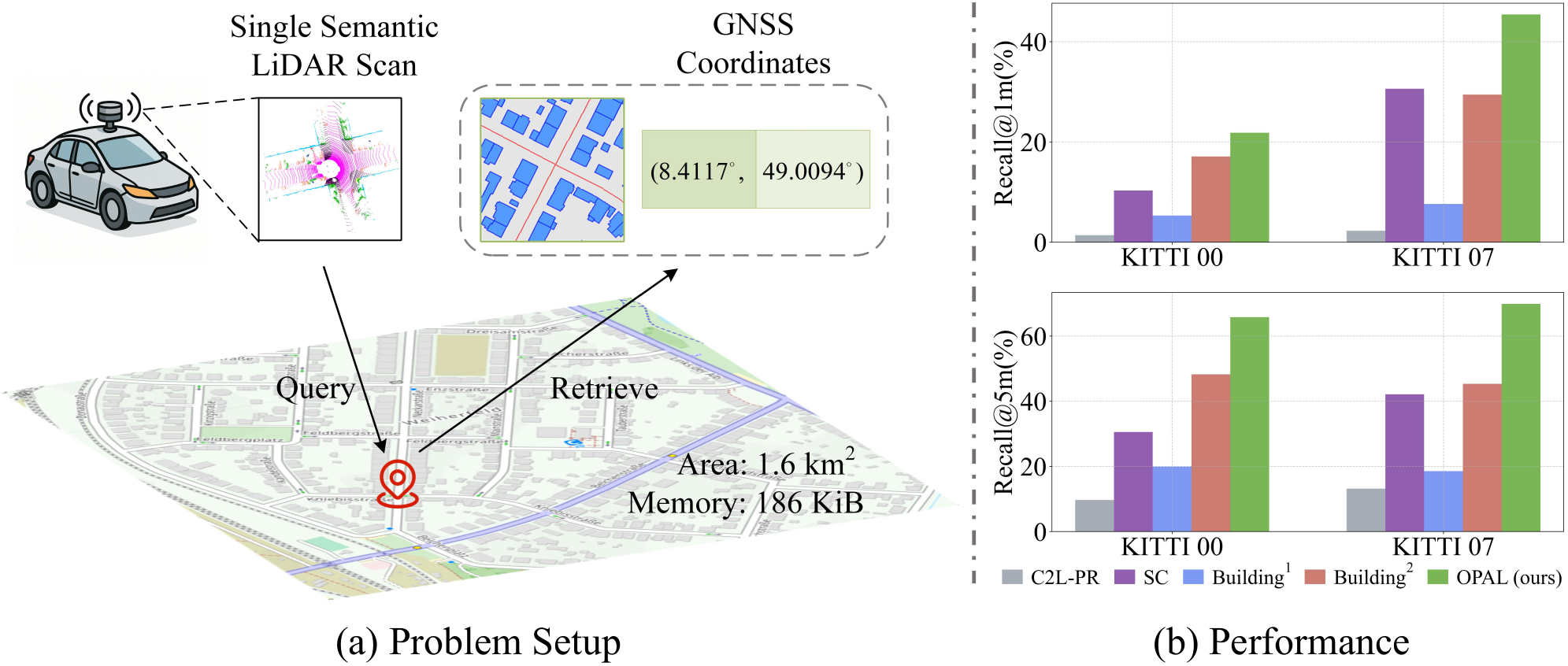}  %
    \caption{(a) Point cloud-to-OpenStreetMap (P2O) place recognition estimates the geographic location of a LiDAR scan by matching the semantic point cloud to geo-referenced OpenStreetMap tiles. (b) shows the evaluation results on KITTI~\citep{geiger2012we} dataset.}

  \label{fig:P2OPR}
  \vspace{-6pt}
 \end{figure}

\makeatletter
\renewcommand{\@makefntext}[1]{\noindent #1} 
\makeatother

\footnotetext[0]{\textsuperscript{*} Equal contribution.\quad\textsuperscript{\textdagger} Corresponding author.}

\begin{abstract}
    LiDAR place recognition is a critical capability for autonomous navigation and cross-modal localization in large-scale outdoor environments. Existing approaches predominantly depend on pre-built 3D dense maps or aerial imagery, which impose significant storage overhead and lack real-time adaptability. In this paper, we propose OPAL, a novel framework for LiDAR place recognition that leverages OpenStreetMap (OSM) as a lightweight and up-to-date prior. Our key innovation lies in bridging the domain disparity between sparse LiDAR scans and structured OSM data through two carefully designed components. First, a cross-modal visibility mask that identifies observable regions from both modalities to guide feature alignment. Second, an adaptive radial fusion module that dynamically consolidates radial features into discriminative global descriptors. Extensive experiments on KITTI and KITTI-360 datasets demonstrate OPAL’s superiority, achieving 15.98\% higher recall at 1\text{m} threshold for top-1 retrieved matches, along with 12$\times$ faster inference speed compared to the state-of-the-art approach. Code and data are publicly available at \url{https://github.com/kang-1-2-3/OPAL}.
\end{abstract}

\section{Introduction}

Accurate and reliable localization is crucial for autonomous vehicles and robots operating in large-scale urban environments, where GNSS signals are often degraded or blocked due to structural obstructions. Place recognition addresses this need by retrieving the most likely location from a reference database, based on a query that reflects the robot's current perception. Compared to image-based place recognition methods~\citep{arandjelovic2016netvlad, hausler2021patch, zhu2023r2former}, which are sensitive to photometric variations caused by changing weather and seasons~\citep{yu2024gv}, LiDAR point clouds maintain robustness under varying illumination and meteorological conditions. Moreover, point cloud offers precise depth measurements and rich geometric detail, making them effective for accurate localization in outdoor environments~\citep{xia2021vpc,chen2020overlapnet}. 

Most existing point cloud-based place recognition methods rely on pre-built 3D maps  ~\citep{uy2018pointnetvlad,liu2019lpd,xia2021soe,luo2024bevplace++} or satellite images~\cite {tang2021get} as reference database. However, constructing a city-scale point cloud map is both time-consuming and costly to maintain, while storage demands remain prohibitively high for large-scale deployments. Although aerial images are more compact than 3D point cloud maps, they are still expensive to capture, generally not free, and heavy to store at high resolution. Moreover, they are highly sensitive to weather, seasonal changes, and lighting conditions. In contrast, OSM provides a globally accessible, compact geospatial database comprising infrastructure, architectural elements, points of interest, land-use classifications and other stationary urban features~\citep{fan2014quality}. Remarkably, it is extremely storage-efficient: only 186\,KiB (as shown in Fig.\ref{fig:P2OPR}(a)), compared with 19.61\,MiB for a bird's-eye view (BEV) point cloud image (as in BEVPlace++\citep{luo2024bevplace++}) or 8.22\,GiB for the raw KITTI 00 sequence. Moreover, OSM data
is continuously updated by volunteers and organizations, with weekly snapshots released. Its timely and rich geometric primitives and semantic elements enable reliable place recognition, mirroring human navigation’s use of spatial and semantic cues~\citep{sarlin2023orienternet,liao2024osmloc}.
\citet{cho2022openstreetmap} first developed a place recognition descriptor for P2O place recognition by calculating the shortest distances to building structures at fixed angular intervals around the sensor.
\citet{lee2024autonomous} proposed a learning-based place recognition method and integrated it into simultaneous localization and mapping (SLAM), while it requires an accurate orientation prior for initialization.
Overall, current single-frame P2O place recognition methods are still limited in accuracy, robustness and efficiency.

In this paper, we present OPAL, a novel P2O place recognition framework that achieves meter-level localization accuracy using a single LiDAR scan, while maintaining real-time computational performance. The OPAL pipeline begins by projecting the query point cloud and OSM data into BEV representation, generating the visibility mask as an additional input to alleviate viewpoint disparity. A Siamese convolutional neural network (CNN) processes these polar representations to extract local feature maps. The adaptive radial fusion (ARF) module then dynamically weights radial-wise features based on their contextual importance, enabling optimized feature aggregation across varying distances and robustness to viewpoint change. Experiments on KITTI and KITTI-360~\citep{liao2022kitti} datasets demonstrate that our method significantly outperforms both hand-crafted and learning-based baselines across various environments. The main contributions include:
\begin{enumerate}
\item We propose a novel pipeline for P2O place recognition. Compared to existing methods, our approach substantially improves accuracy, robustness, and computational efficiency.
\item We introduce visibility mask to resolve the viewpoint disparity between cross-modal inputs. The visibility mask significantly improves cross-modality feature alignment by focusing on mutually visible regions and ignoring modality-specific occlusions.
\item We propose the ARF module to dynamically fuse radial features into the global descriptor. This adaptive strategy preserves geometric structure while maintaining real-time efficiency.
\end{enumerate}

\section{Related Work}\label{sec:Related Works}

We review point cloud place recognition research through two perspectives: uni-modal point cloud place recognition approaches and cross-modal approaches that bridge different sensor domains.

\textbf{Uni-modal point cloud place recognition.}
Early breakthroughs in point cloud-to-point cloud place recognition were led by PointNetVLAD~\citep{uy2018pointnetvlad}, which combined PointNet~\citep{qi2017pointnet} with the NetVLAD~\citep{uy2018pointnetvlad} aggregation layer to produce global descriptors from raw point clouds. 
Transformer-based architectures have also been explored for capturing long-range dependencies and contextual semantics~\citep{fan2022svt, zhang2022rank, ma2022overlaptransformer}, leveraging attention mechanisms to improve feature expressiveness. MinkLoc3D~\citep{komorowski2021minkloc3d} employed a voxel-FPN architecture with generalized mean pooling (GeM) for compact global descriptors. Recently, CASSPR~\citep{xia2023casspr} proposed a hybrid voxel-point dual-branch framework using hierarchical cross-attention to effectively fuse multi-level features, significantly boosting performance on sparse single-frame scans.
Although these methods leverage the rich spatial information from LiDAR data to achieve strong performance, their scalability is limited by the high cost and maintenance overhead associated with constructing and updating dense, city-scale point cloud maps. These practical limitations pose a major obstacle in consumer-grade applications.

\textbf{Cross-Modal point cloud place recognition.}
For image-to-point (I2P) cloud place recognition,  \citet{cattaneo2020global} and \citet{li2024vxp} established a shared global feature space for feature matching and retrieval. C2L-PR~\citep{xu2024c2l} improved I2P place recognition via modality alignment and orientation voting. For point cloud-to-aerial image place recognition, \citet{tang2021get} proposed a self-supervised localization approach based on 2D occupancy map matching. Beyond I2P place recognition, recent efforts have extended cross-modal localization to natural language queries~\citep{kolmet2022text2pos,xia2024text2loc,xia2024uniloc}. 

OpenStreetMap-based approaches are most related to our method. OpenStreetSLAM~\citep{floros2013openstreetslam} integrated visual odometry with map priors to improve trajectory accuracy, while subsequent methods~\citep{ruchti2015localization, vysotska2016exploiting} focused on road or building structure alignment with OSM data. \citet{suger2017global} introduced a Monte Carlo localization framework that aligns semantic features from LiDAR with the OSM data for outer-urban navigation. \citet{yan2019global} proposed a compact 4-bit descriptor that encoded the street intersections and building gaps for efficient global localization. \citet{bieringer2024analyzing} utilized Level of Detail 3 (LOD3) models for outdoor map-based positioning. Sequential frames generally improve accuracy through spatial consistency, yet single-scan place recognition remains critical in unknown or dynamic environments that lack accurate maps. Besides, methods developed for sequential point cloud localization often struggle in the single-frame setting, where limited observations and the absence of motion constraints significantly hinder performance.
For the single-frame P2O place recognition,~\citet{cho2022openstreetmap} proposed a hand-crafted descriptor by extracting the shortest distance to buildings at fixed angular intervals for cross-modality feature matching, later improved by \citet{li2024lidar} with directional boundary features. However, these methods exhibit strong dependence on building structures, limiting practical applicability. 
Although \citet{lee2024autonomous} introduced learning-based descriptors, their method requires IMU-based orientation priors for initialization.  To summarize, existing solutions suffer from three key limitations: reliance on sequential inputs, limited accuracy and robustness, and inefficient descriptor generation. To address these challenges, we propose OPAL, a learning-based P2O place recognition framework that unifies geometric and topological cues to enable accurate, robust, and generalizable localization across diverse environments.

\vspace{-6pt}
\section{Methodology}\label{sec:Methodology}

The P2O place recognition task aims to localize a query LiDAR point cloud $\mathcal{P} \in \mathbb{R}^{N \times 3}$ by matching it against a geo-referenced OSM database $\mathbb{O}$, where each of the $N$ points in $\mathcal{P}$ is represented by a 3D Cartesian coordinate $(x, y, z)$. Since our approach incorporates cross-modality data, $\mathcal{P}$ and  $\mathbb{O}$ require pre-processing before being fed into the framework. $\mathcal{P}$ is first enhanced by concatenating per-point semantic labels as $\mathcal{P}' \in \mathbb{R}^{N \times 4}$. The original OSM data $\mathbb{O}$ is stored in structured format and represents various entities, including areas, ways, and nodes, in geographic coordinate system. Following the OrienterNet~\citep{sarlin2023orienternet}, we rasterize the areas, ways, and nodes into a 3-channel grid map with a fixed sampling distance $\Delta_o$ in local 2D East-North coordinate system. From this projected map, we densely sample 
$m$ map tiles $\mathbb{O}={\{\mathcal{O}_i\}}_{i=1}^m$ along the ego-vehicle trajectory to construct the OSM database, where each tile $\mathcal{O}_i$ corresponds to an $H \times W$ meters region centered at geographic coordinates $(lat_i,lon_i)$. Details of the OSM tile database are given in Appendix~\ref{appendix: pre-processing}.

Fig.~\ref{fig:workflow} illustrates OPAL's pipeline. The pipeline begins by computing visibility masks to resolve occlusion patterns caused by viewpoint disparities (Sec.~\ref{sec:Visibility Mask}). Next, a Siamese polar CNN architecture is employed to extract deep feature maps from both modalities (Sec.~\ref{sec:feature extraction}), which are subsequently aggregated into compact global descriptors through the proposed ARF module (Sec.~\ref{sec: active radial fusoin}). %

\begin{figure*}[!t]
  \centering
  \includegraphics[width=\linewidth]{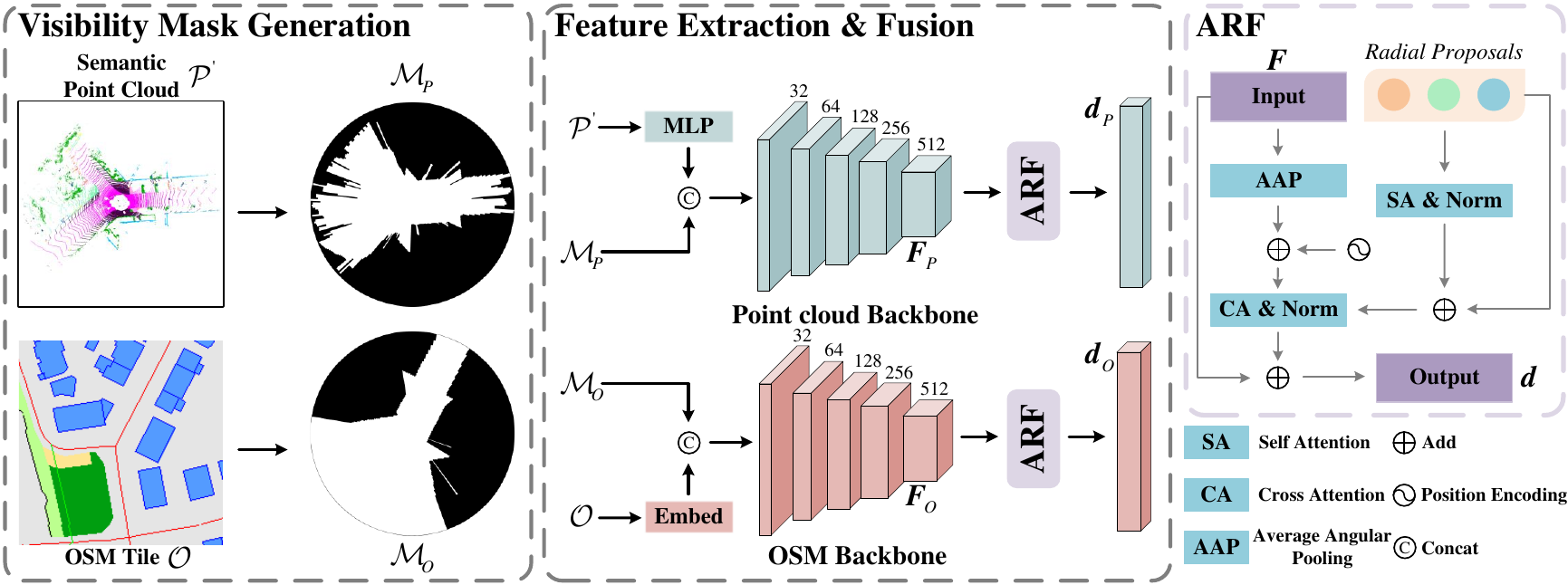}
  \captionsetup{justification=justified,singlelinecheck=false}
  \caption{Overview of proposed OPAL. Given a semantic point cloud frame $\mathcal{P}'$ and OSM tile $\mathcal{O}$, OPAL computes visibility masks to bridge the occlusion difference, then extracts polar BEV features via a Siamese encoder, and lastly generates discriminative descriptors using ARF for place retrieval.}  %
  \label{fig:workflow}
  \vspace{-4pt}
\end{figure*}

\subsection{Visibility Mask Generation}\label{sec:Visibility Mask}

The concept of visibility alignment originates from prior work in occlusion handling for image matching~\citep{fan2023occ}, visual localization~\citep{tang2021get,sarlin2023orienternet}, and 3D building reconstruction~\citep{wysocki2023scan2lod3}. In the P2O place recognition, this challenge remains significant due to the modality gap between LiDAR scans and OSM data. Effective visibility handling becomes crucial for robust cross-modal matching.

To address this issue, we compute visibility masks $\mathcal{M}$ for both point cloud and OSM data to resolve occlusion discrepancies. Given a point cloud frame $\mathcal{P}' \in \mathbb{R}^{N \times 4}$, we first project it onto a polar BEV grid with $U$ radial rings and $V$ angular sectors, assigning points within each cell $(u,v)$. The radial and angular resolutions are given by $\Delta_r = \frac{L}{U}$ and $\Delta_s = \frac{2\pi}{V}$, where $L$ is defined as the maximum valid range of LiDAR.  For each polar cell $(u, v)$, the corresponding radial distance $r_{u,v}$ and azimuth angle $\phi_{u,v}$ are defined as: 
\begin{equation} 
\begin{aligned} 
& r_{u,v} = (u + 0.5)\Delta_r, && u \in \{0,\dots,U-1\}, \\
& \phi_{u,v} = (v + 0.5)\Delta_s, && v \in \{0,\dots,V-1\}. 
\end{aligned} 
\label{eq:polar_grid} 
\end{equation}
Through ray casting, cells are classified as visible $\mathcal{M}_P(u,v)=1$ if they lie within the line-of-sight before the last measured return in a sector. Conversely, cells are marked as occluded $\mathcal{M}_P(u,v)=0$ if they are behind the last valid range return :
\begin{align}
\mathcal{M}_P(u,v) = 
\begin{cases} 
1, &   r_{u,v} \leq {\mathrm{max}(}r_{[:,v]}\text{)} \\
0, &  \text{otherwise} 
\end{cases}
\end{align}
where  $ {\mathrm{max}(}r_{[:,v]}\text{)}$ is range of the last valid return in $v$-th sector.

For the OSM tile $\mathcal{O} \in \mathbb{R}^{H \times W \times 3}$ defined in the 2D Cartesian coordinate system, we convert it into a polar BEV grid with $U$ rings and $V$  sectors, as in the point cloud branch. For each polar cell $(u, v)$, the corresponding Cartesian coordinates $(x_{u,v}, y_{u,v})$ are computed via:
\begin{equation}
\label{eq:bilinear} 
x_{u,v} = r_{u,v} \cos\phi_{u,v}, \quad y_{u,v} = r_{u,v} \sin\phi_{u,v}, 
\end{equation}

and then the polar representation is obtained by bilinear interpolation of $\mathcal{O}$ at $(x_{u,v}, y_{u,v})$.

As OSM data lacks explicit range measurements, visibility estimation relies on semantic cues. Here, we select the ``building" elements from the area channel as occluders, owing to their vertical extent and structural continuity, which consistently obstruct sensor visibility in both urban and suburban environments. Through ray-casting in each sector, cells are classified as occluded $\mathcal{M}_O(u,v)=0$ if they lie further than the nearest ``building" element; otherwise, they are classified as visible $\mathcal{M}_O(u,v)=1$.  The process is formatted as:
\begin{equation}
\mathcal{M}_O(u,v) = 
\begin{cases}
0, & u > \text{min(}u'\text{)} \text{\: if  \:} \exists \mathcal{O}(u',v) = \text{``building"}  \\
1, & \text{otherwise}
\end{cases}
\label{eq:visibility_mask}
\end{equation}

where $\mathcal{O}(u',v)$ is the building elements in $v$-th sector.

\textbf{Remark.} Unlike prior approaches~\citep{tang2021get,sarlin2023orienternet} that estimate the visible or confidential mask with a neural network, our visibility mask generation is fully deterministic and leverages the complementary strengths of both modalities. For LiDAR, valid range measurements directly yield visibility masks. For OSM, which lacks range information, we approximate the visibility mask using building masks, as buildings are the primary occluders in urban scenes. By eliminating the approximation errors and training overhead of learned visibility estimation, our method preserves geometric consistency across modalities while ensuring computational efficiency.

\subsection{Feature Extraction}\label{sec:feature extraction}

As shown in Fig.~\ref{fig:workflow}, our feature extraction pipeline processes both modalities through parallel yet symmetric branches. The augmented point cloud $\mathcal{P}' \in \mathbb{R}^{N \times 4}$ is first passed through a lightweight multilayer perceptron (MLP) to generate $C_{pem}$-dimensional per-point features. These features are then splatted onto the polar BEV grid, where grid-wise features are aggregated using max pooling, resulting in a dense feature map $\boldsymbol{F}_{P} \in \mathbb{R}^{U \times V \times C_{pem}}$. This representation is concatenated with the visibility mask $\mathcal{M}_P \in \mathbb{R}^{U \times V \times 1}$, and processed by the encoder of PolarNet~\citep{zhang2020polarnet}, yielding the local feature map $\boldsymbol{F}_{P}'\in \mathbb{R}^{Z \times T \times C}$. 

For the OSM branch, we embed each channel of the rasterized map tile $\mathcal{O} \in \mathbb{R}^{H \times W \times 3} $ into a $C_{oem}$-channel feature, generating dense semantic feature $\boldsymbol{F}_{O} \in \mathbb{R}^{H \times W \times (3 \times C_{oem})}$ in the Cartesian coordinate system. Then $\boldsymbol{F}_{O}$ is transformed into polar BEV feature map via bilinear sampling with Eq.~(\ref{eq:bilinear}), and concatenated with the visibility mask $\mathcal{M}_O \in \mathbb{R}^{U \times V \times 1}$ to form a visibility-aware input. Finally, the OSM features are processed through a separate PolarNet~\citep{zhang2020polarnet} encoder (with weights independent of the point cloud branch) to produce the final feature map $\boldsymbol{F}_O' \in \mathbb{R}^{Z \times T \times C}$.

\subsection{Adaptive Radial Fusion}\label{sec: active radial fusoin}
Given the extracted local features  $\boldsymbol{F} \in \mathbb{R}^{Z \times T\times C}$ from the point cloud and OSM tile, the next step is to aggregate them into a global descriptor for efficient retrieval. This global descriptor is expected to be both representative and robust to orientation variations.
Existing solutions entail critical trade-offs: frequency-domain methods~\citep{xu2023ring++,lu2023deepring} and range projection approach~\citep{ma2022overlaptransformer} sacrifice spatial relationships for rotation invariance, while sampling-based approaches like BEVPlace++~\citep{luo2024bevplace++}  suffer from high computational overhead. These limitations hinder cross-modal place recognition by either compromising geometric fidelity or reducing system efficiency.

To preserve geometric completeness and ensure rotation robustness, we introduce the ARF module (shown in the last column of Fig.~\ref{fig:workflow}). The module first extracts radial features through angular average pooling (AAP):
\begin{equation}
    \boldsymbol{F}_r = \frac{1}{T}\sum_{t=1}^T\boldsymbol{F}[:,t,:] + \boldsymbol{E}_{re}, 
\end{equation}
where $\boldsymbol{F}_r$ represents the radially compressed features, and $\boldsymbol{E}_{re} \in \mathbb{R}^{Z \times C}$ encodes the ring-order information using cosine position encoding~\citep{vaswani2017attention}. Building upon the radial-wise features, we introduce trainable \textit{radial proposals} $\boldsymbol{Q} \in \mathbb{R}^{Z \times C}$ as in ~\citep{carion2020end,ali2024boq},  to adaptively track and fuse the radial-wise features based on significance. These \textit{radial proposals} are implemented as trainable model parameters that undergo a two-stage attention refinement process. First, inter-proposal communication is enhanced with a self-attention module: 
\begin{equation}\label{eq.sa} 
\boldsymbol{Q}'=\mathrm{softmax}(\frac{\boldsymbol{Q}\boldsymbol{Q}^{\top}}{\sqrt{C}})\boldsymbol{Q}.
\end{equation}
This enables the proposals to capture global contextual awareness while suppressing redundant correlations. The refined proposals then selectively aggregate information from the radial features via a cross-attention mechanism:
\begin{equation}
    \boldsymbol{F}_r' = \mathrm{softmax}(\frac{\boldsymbol{Q}'\boldsymbol{F}_r^{\top}}{\sqrt{C}})\boldsymbol{F}_r.
\end{equation}
This attention mechanism dynamically weights each radial feature based on its geometric salience for place recognition, allowing the model to focus on discriminative spatial patterns while maintaining robustness to orientation variations.
The refined radial features   $\boldsymbol{F}_r'$ are then combined with the original  $\boldsymbol{F}_r$ via a residual connection and transformed into the global descriptor $\boldsymbol{d}$ finally:
\begin{equation}
    \boldsymbol{d} = \boldsymbol{W}(Flatten(\boldsymbol{F}_r + \boldsymbol{F}_r')),
\end{equation}
where $Flatten(\cdot)$ denotes the operation that flattens the radial-wise features into a one-dimensional vector, and $\boldsymbol{W}$ is a fully connected layer that projects this vector to the global descriptor $\boldsymbol{d}$.

\textbf{Remark.} The key innovation lies in combining LiDAR's native radial geometry with learned attention. AAP preserves sensor's scanning pattern while ensuring rotation robustness, and the trainable radial proposals adaptively weight features based on their importance. The proposed ARF surpasses fixed aggregation methods without compromising computational efficiency.

\section{Experiments}
\subsection{Experimental Setup} 

 We validate the proposed method on two public datasets: KITTI~\citep{geiger2012we} and KITTI-360~\citep{liao2022kitti}. The corresponding OSM data is collected from the OpenStreetMap official website\footnote{\href{https://www.openstreetmap.org/}{https://www.openstreetmap.org/}}.  Our model is trained exclusively on the KITTI dataset and evaluated in a zero-shot setting on KITTI-360 to assess its generalization capability. For evaluation, we adopt the standard Recall@$K\text{m}$ metric, which measures the proportion of queries whose top-1 retrieved match falls within $K$-meters of the ground-truth (GT) location. To prevent overlap between the training and validation sets, sequences 00 and 07 from KITTI, and sequences 00, 05, 06, and 09 from KITTI-360 are used for evaluation. Additional details about the train/test splits are provided in Appendix.~\ref{appendix: dataset}. 
 
 \textbf{Baseline methods.} Due to the limited attention of P2O place recognition research, we adapt three place recognition baselines for comparison: a) \underline{Building}~\citep{cho2022openstreetmap}:  the pioneering P2O place recognition method that first utilizes a global key for fast matching, followed by a fine-grained descriptor for precise localization.   We re-implement both the original two-stage version (Building$^2$) and a simplified one-stage variant (Building$^1$) for comparison. 
b) \underline{SC}~\citep{kim2018scan}:  a widely-used point cloud-to-point cloud place recognition method based on hand-crafted descriptors. Following \citet{li2024lidar}, we extract building points from both LiDAR scans and OSM data to facilitate descriptor extraction and matching.
c) \underline{C2L-PR}~\citep{xu2024c2l}: A hybrid framework for image-to-point cloud place recognition. C2L-PR first extracts hand-crafted features from point cloud semantics (road, parking, sidewalk, other-ground, building, fence, other-structure, vegetation, terrain) and OSM data (building, parking, grass, forest, fence, wall, road), then learns the descriptors via an embedding network.  We re-trained the modified C2L-PR with the same setting as OPAL for fair comparison.

\subsection{Place Recognition Results} \label{sec:Results}

\setlength{\intextsep}{2pt}  
\begin{wraptable}{r}{0.5\linewidth} 
\centering
\customsize  
\caption{Recall@$K$ of top-1 retrieved results on the KITTI dataset.}  
\setlength{\tabcolsep}{2pt}
\label{tab:kitti}
\begin{tabular}{@{}c|ccc|ccc@{}}
\toprule
\multirow{2}{*}{Method} & \multicolumn{3}{c|}{Seq 00} & \multicolumn{3}{c}{Seq 07} \\
\cmidrule(lr){2-4} \cmidrule(lr){5-7}
& R@1 & R@5 & R@10 & R@1 & R@5 & R@10 \\
\midrule
SC~\citep{kim2018scan} & 10.31 & 30.54 & 31.16 & 30.61 & 42.42 & 43.42 \\
Building$^1$~\citep{cho2022openstreetmap} & 5.31 & 19.89 & 20.52 & 7.63 & 18.53 & 19.35 \\
Building$^2$~\citep{cho2022openstreetmap} & 17.09 & 48.23 & 48.93 & 29.43 & 45.32 & 45.78 \\
C2L-PR~\citep{xu2024c2l} & 1.39 & 9.69 & 12.20 & 2.27 & 13.17 & 17.26 \\
\midrule
OPAL      & \textbf{21.82} & \textbf{65.78} & \textbf{66.40} & \textbf{45.41} & \textbf{69.85} & \textbf{70.30} \\
OPAL-Rot  & 21.49 & 66.46 & 67.14 & 46.14 & 70.12 & 70.30 \\
\bottomrule
\end{tabular}
\end{wraptable}

\begin{figure*}[t]

  \centering
  \includegraphics[width=\linewidth]{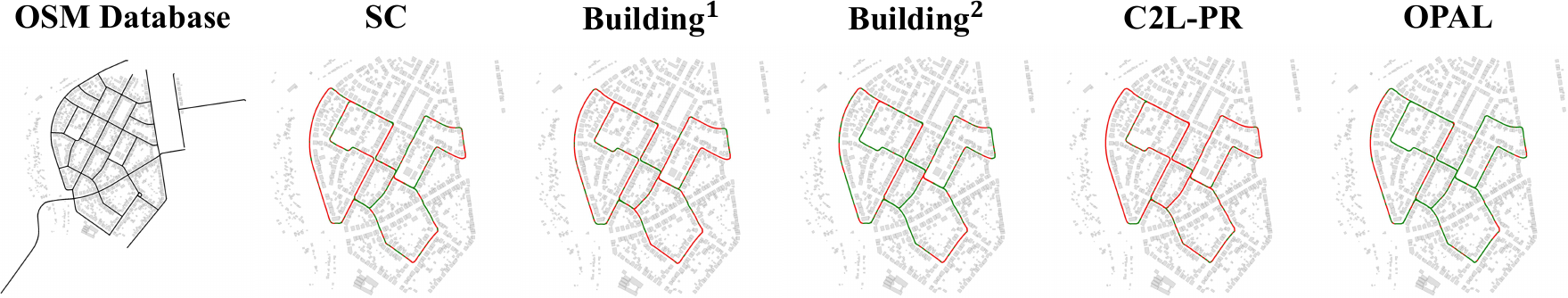}
\captionsetup{justification=justified,singlelinecheck=false}
  \caption{Top-1 retrieved results @5m threshold on the 00 sequence of the KITTI dataset. Black points \textcolor{black}{$\bullet$} denote OSM tile locations, while red \textcolor{red}{$\bullet$} and green \textcolor{green}{$\bullet$} indicate the wrong and corrected retrieved results, respectively.}
  \label{fig:traj}
  \vspace{1mm}
  
  \centering
  \includegraphics[width=\linewidth]{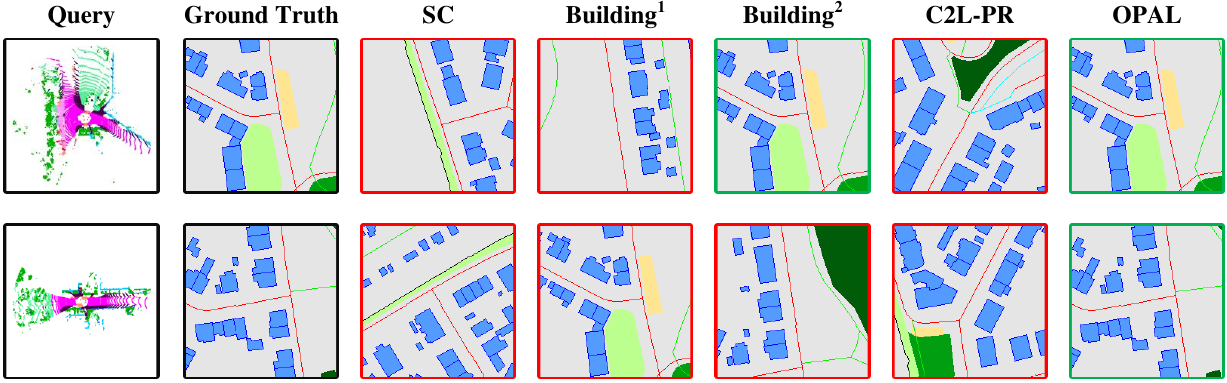}
  \captionsetup{justification=justified,singlelinecheck=false}
  \caption{Examples of LiDAR queries and their top-1 retrieved matches on KITTI. Red rectangles \( \raisebox{0.5ex}{\textcolor{red}{\fbox{}}} \) represent wrong retrieved results and green \( \raisebox{0.5ex}{\textcolor{green}{\fbox{}}} \) represent correct retrieved results. Legends for point cloud and OSM tile are shown in the Appendix~\ref {appendix: pre-processing}.}
  \label{fig:qualitative}

  \vspace{-9mm}
\end{figure*}

\textbf{KITTI.} As shown in Tab.~\ref{tab:kitti}, OPAL achieves superior place recognition performance on KITTI sequences 00 and 07, outperforming all baseline methods by a large margin. Specifically, it achieves significant improvements of 4.73\%, 17.55\%, and 17.47\% in R@1m/5m/10m metrics on sequence 00, and 15.98\%, 24.53\%, and 24.52\% on sequence 07 compared to the state-of-the-art method, Building$^2$~\citep{cho2022openstreetmap}. Fig.~\ref{fig:traj} illustrates OPAL's accurate localization results along the trajectory across diverse environments,  highlighting the robust performance of our method under various conditions. Fig.~\ref{fig:qualitative} shows the qualitative results of the baseline methods and our OPAL. These quantitative and qualitative results demonstrate OPAL’s effectiveness in enhancing localization accuracy and robustness.

\textbf{Robust to rotation.} To assess rotational robustness, we apply random z-axis rotations uniformly sampled from $[0, 2\pi]$ to each query point cloud to simulate view change. As shown in the last row of Tab.~\ref{tab:kitti}, OPAL remains robust under these transformations.

\begin{table*}[hbt!]

    \centering
    \customsize
    \caption{Recall@$K$m of top-1 retrieved results on the KITTI-360 dataset.}
    \label{tab:kitti360}
    \renewcommand{\arraystretch}{1.0}
    \setlength{\tabcolsep}{0.37em}
    \begin{tabular}{@{}c|ccc|ccc|ccc|ccc@{}}
        \toprule
        \multirow{2}{*}{Method} 
        & \multicolumn{3}{c|}{Seq 00} 
        & \multicolumn{3}{c|}{Seq 05} 
        & \multicolumn{3}{c|}{Seq 06} 
        & \multicolumn{3}{c}{Seq 09} \\
        \cmidrule(lr){2-4} \cmidrule(lr){5-7} \cmidrule(lr){8-10} \cmidrule(lr){11-13}
        & R@1 & R@5 & R@10 & R@1 & R@5 & R@10 & R@1 & R@5 & R@10 & R@1 & R@5 & R@10 \\
        \midrule
        SC~\citep{kim2018scan} & 15.14 & 39.61 & 40.66 & 3.69 & 16.69 & 17.18 & 4.14 & 14.22 & 14.59 & 13.92 & 27.45 & 28.07\\
        Building$^1$~\citep{cho2022openstreetmap}  & 5.22  & 15.76 & 17.07 & 0.87  & 3.91 & 4.61 & 0.60  & 3.18 & 3.79 & 4.21 & 12.16 & 13.27 \\
        Building$^2$~\citep{cho2022openstreetmap}  & \textbf{17.12} & \textbf{49.61} & \textbf{51.29} & 4.23 & 15.94 & 16.29 & 3.00 & 12.82 & 13.39 & 18.28 & 41.92 & 42.64 \\
        C2L-PR~\citep{xu2024c2l}  & 1.70 & 8.23 & 10.93 & 0.81 & 4.72 & 6.36 & 0.45 & 3.35 & 4.50 & 1.69 & 9.59 & 12.45 \\ 
        \midrule
        OPAL   & 14.92  & 42.82 & 44.18 & \textbf{7.74} & \textbf{30.49} & \textbf{31.55} & \textbf{7.71} & \textbf{36.38} & \textbf{37.54} & \textbf{27.89} & \textbf{60.96} & \textbf{61.92} \\
        \bottomrule
    \end{tabular}
    \vspace{2mm}
\end{table*}

\begin{figure*}[hbt!]
  \centering
  \includegraphics[width=\linewidth]{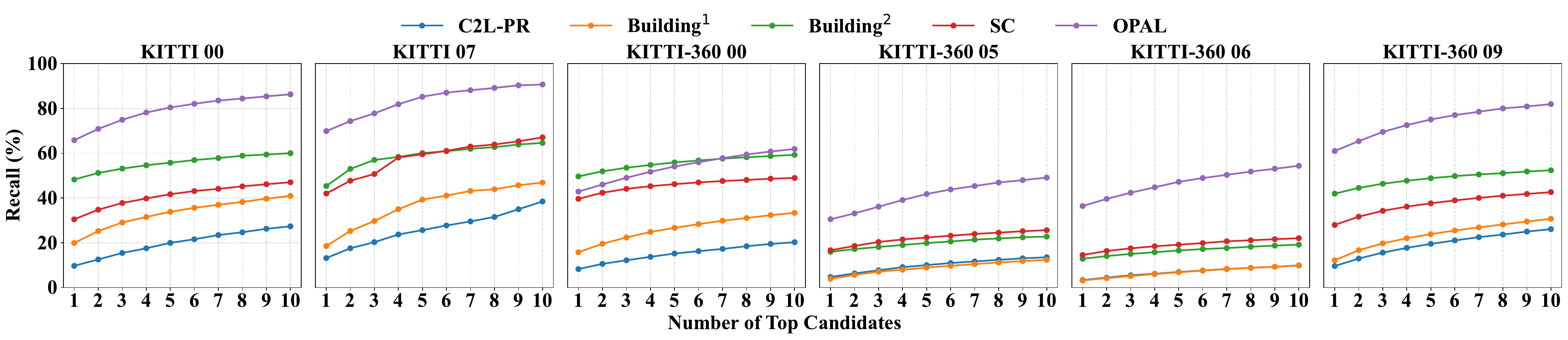}
  \captionsetup{justification=justified,singlelinecheck=false}
  \caption{Recall curves @5m of top-N candidates on the KITTI and KITTI-360 datasets.}
  \label{fig:recall}

\end{figure*}
\textbf{Zero-shot generalization on KITTI-360.} As shown in Tab.~\ref{tab:kitti360}, while OPAL shows slightly reduced performance in building-dominated urban environments (sequence 00) compared to the Building$^2$~\citep{cho2022openstreetmap}, it achieves significant improvements over baselines in other sequences 05, 06, 09, with performance gains of 14.55\%, 23.56\%, and 19.04\% at R@5m, respectively. Fig.~\ref{fig:recall} presents the recall curves of top candidates within a 5-meter threshold on the KITTI and KITTI-360 datasets, where OPAL consistently outperforms baseline methods across diverse scenes. These results highlight OPAL’s strong generalization capability across diverse environments.

\begin{wraptable}{r}{0.48\linewidth}
    \centering
    \customsize
    \renewcommand{\arraystretch}{1.0}
    \setlength{\tabcolsep}{2pt}
    \caption{Descriptor generation runtime (ms).}
    \label{table:runtime}
    \begin{tabular}{@{}c|ccc@{}}
        \toprule
        Method & Point Cloud & OSM Tile & Total \\
        \midrule
        SC~\citep{kim2018scan} & 31.54 & 16.68 & 48.22 \\
        Building~\citep{cho2022openstreetmap} & 29.86 & 54.87 & 84.73 \\
        C2L-PR~\citep{xu2024c2l} & 219.08 & 316.71 & 535.79 \\
        OPAL & \textbf{1.91} & \textbf{5.14} & \textbf{7.05} \\
        \bottomrule
    \end{tabular}
\end{wraptable}
\textbf{Runtime performance.} We evaluate our method on a desktop with an Intel i9-13900K CPU and NVIDIA RTX 4090 GPU and report the results in Tab.~\ref{table:runtime}. Our OPAL achieves high efficiency, processing point clouds in 1.91\,ms and OSM tiles in 5.14\,ms, resulting in a total runtime of only 7.05\,ms. This corresponds to a throughput exceeding 140 FPS for descriptor generation, enabling deployment in time-sensitive applications.

\subsection{Ablation Study} \label{sec:Ablation Study} 
We conduct ablation studies on KITTI sequence 00 to evaluate the impact of three components: the visibility mask, ARF module, and the effect of semantic labels in the point cloud.

\begin{wraptable}{r}{0.41\linewidth}
    \centering
    \customsize
    \setlength{\tabcolsep}{2pt}
    \renewcommand{\arraystretch}{1.0}
    \caption{Ablation study on visibility mask and ARF module.}
    \label{tab:ablation_vmfa}
    \begin{tabular}{c|cc|ccc}
        \toprule
        ID & VM & FA & R@1 & R@5 & R@10 \\
        \midrule 
        {[A]} &            & GAP      & 3.79  & 22.46 & 24.82 \\
        {[B]} &            & ARF      & 21.03 & 62.21 & 63.27 \\
        {[C]} & \checkmark & GAP      & 5.84  & 17.95 & 19.18 \\
        {[D]} & \checkmark & GeM      & 1.56  & 9.36  & 10.52 \\
        {[E]} & \checkmark & NetVLAD  & 6.39  & 30.70 & 32.26 \\
        {[F]} & \checkmark & MixVPR   & 1.78  & 7.69  & 8.17 \\
        {[G]} & \checkmark & AAP      & 17.68 & 51.62 & 52.17 \\
        {[H]} & \checkmark & ARF      & \textbf{21.82} & \textbf{65.78} & \textbf{66.40} \\
        \bottomrule
    \end{tabular}
    \caption*{\footnotesize VM: visibility mask; FA: feature aggregation.}
\end{wraptable}
\textbf{Visibility mask and ARF.} 
Tab.~\ref{tab:ablation_vmfa} presents a systematic comparison of seven architectural variants, all trained from scratch. Variant $[A]$ represents a simple baseline with PolarNet~\citep{zhang2020polarnet} for feature extraction and global average pooling (GAP) for feature aggregation, without the visibility mask and ARF module. $[B]$ introduces ARF alone, yielding notable performance gains. For variants $[A]$ and $[B]$, the visibility mask is removed during both training and evaluation. With the visibility mask, variants $[C]$ and $[D]$ adopt GAP and GeM, while $[E]$ and $[F]$ use NetVLAD~\citep{arandjelovic2016netvlad} and MixVPR~\citep{ali2023mixvpr}, respectively. All variants, however, show limited performance in the P2O place recognition task, mainly because global pooling-based aggregators discard critical local features, and NetVLAD/MixVPR are tailored for front-view images rather than BEV point clouds, making them rotation-sensitive and less effective in our setting. $[G]$ employs the average angular pooling (AAP) for feature aggregation and achieves moderate performance. Finally, variant $[H]$ combines both the visibility mask and ARF module, achieving the best performance, confirming the effectiveness of their joint contribution.

\begin{wraptable}{r}{0.48\linewidth}
    \centering
    \customsize
    \renewcommand{\arraystretch}{1.0}
    \setlength{\tabcolsep}{2pt}
    \caption{Effect of semantic label in point cloud.}
    \label{tab:ablation_semantic_labels}
    \begin{tabular}{@{}c|ccc@{}}
        \toprule
        Semantic Label & R@1 & R@5 & R@10 \\
        \midrule
        Rangenet++~\citep{milioto2019rangenet++} & 18.92 & 59.08 & 60.30 \\
        Cylinder3D~\citep{zhu2021cylindrical} & 21.82 & 65.78 & 66.40 \\
        Ground Truth~\citep{behley2019semantickitti} & 25.37 & 73.99 & 74.68 \\
        \bottomrule
    \end{tabular}
\end{wraptable}
\textbf{Effect of point cloud semantic label accuracy.} Tab.~\ref{tab:ablation_semantic_labels} illustrates the impact of semantic labels on P2O place recognition performance. With the ground truth labels~\citep{behley2019semantickitti}, our OPAL achieves the best results (74.68\% R@10m), followed by predicted labels from  Cylinder3D~\citep{zhu2021cylindrical} (66.4\%) and Rangenet++~\citep{milioto2019rangenet++} (60.30\%). The 14.38\% performance gap between Rangenet++ and GT annotations underscores the potential reliance of the framework’s accuracy on the precision of semantic labels within the point cloud.

\section{Conclusion}
\label{sec:conclusion}
    In this work, we presented OPAL, a novel single-frame P2O place recognition framework. The proposed method introduces the visibility-aware mask to resolve the cross-modality occlusion, coupled with the adaptive radial fusion module for effectively and robustly global descriptor aggregation. Experiments on the KITTI and KITTI-360 datasets demonstrate that OPAL consistently outperforms state-of-the-art baseline methods across diverse challenging scenarios, significantly improving accuracy and computational efficiency.

\newpage
\section{Limitation}
\begin{figure*}[ht]
\vspace{6pt}
  \centering
  \includegraphics[width=\linewidth]{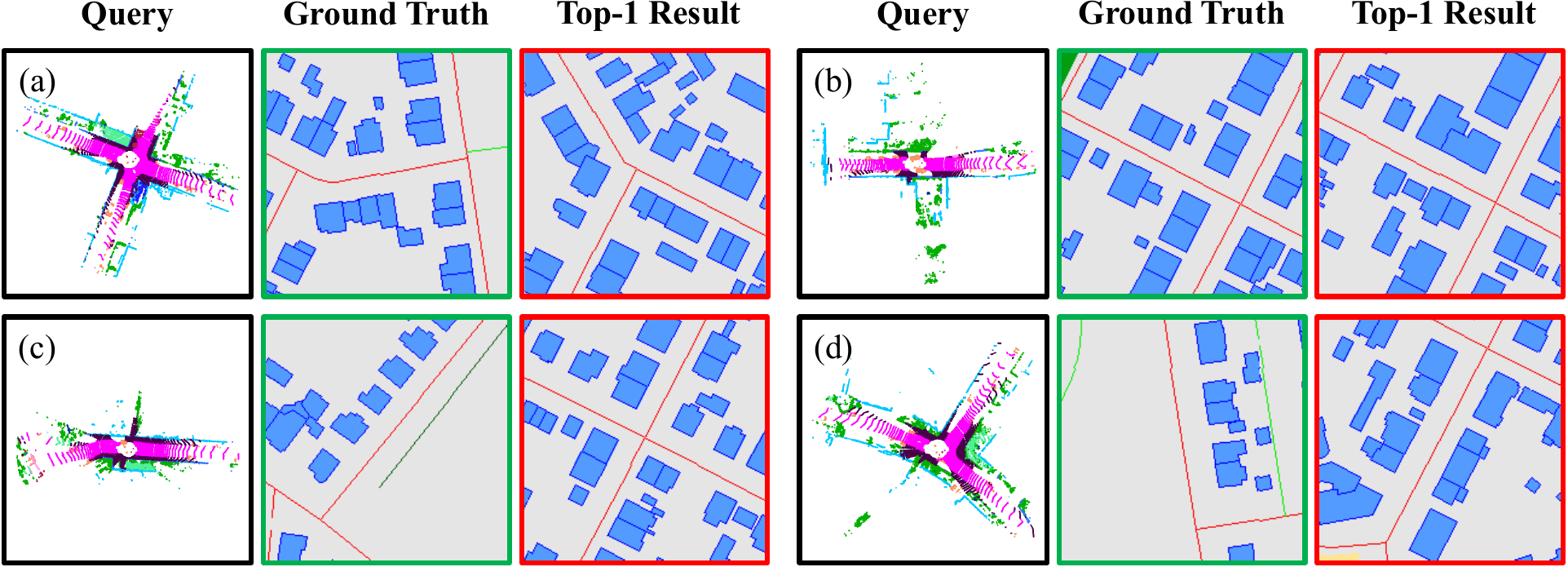}
  \captionsetup{justification=justified,singlelinecheck=false}
  \caption{Failure cases.  The red rectangle \( \raisebox{0.5ex}{\textcolor{red}{\fbox{}}} \) represents the wrong retrieved top-1 result and the green rectangle represents \( \raisebox{0.5ex}{\textcolor{green}{\fbox{}}} \) the GT OSM tile.}   %
  \label{fig:failure_case}
\end{figure*}

The localization accuracy of our OPAL heavily depends on the quality and distinctiveness of the surrounding objects in the point cloud. Fig.~\ref{fig:failure_case} shows some failure cases under different conditions. (a)-(b) show ambiguous scenarios at road crossings with limited distinctive features, leading to top-1 retrieval errors due to the geometric similarity between the retrieved and ground-truth locations. As shown in (c)-(d), cross-modal discrepancies occur when roadside vegetation and buildings detected in LiDAR scans are missed in the OSM data. Furthermore, as discussed in Sec.~\ref{sec:Ablation Study}, the localization accuracy is strongly influenced by the precision of semantic labels assigned to the point cloud. To address these limitations, we plan to: 1) extend OPAL to sequential point cloud-based place recognition, which could leverage temporal and geometric consistency to improve the reliability and accuracy; 2) incorporate orientation priors, text, or images to reduce dependence on point cloud semantic labels.

\bibliography{reference} 

\newpage
\appendix

\section{Complementary Datasets}\label{appendix: dataset}

The KITTI~\citep{geiger2012we} dataset contains LiDAR scans collected from urban driving trajectories in Karlsruhe, with poses provided by integrated GPS/IMU systems. To avoid overlap between training and testing regions, we use sequences 01, 02, 04, 05, 06, and 08 for training, and sequences 00 and 07 for testing. Sequence 03 is excluded due to unavailable GPS information.

The KITTI-360~\citep{liao2022kitti} extends the KITTI dataset with longer suburban routes.  Following the former practice~\citep{cho2022openstreetmap}, synchronized sequences 00, 05, 06, and 09 are utilized for testing. Statistics of query point cloud frames, osm tiles and trajectory length are shown in Tab.~\ref{tab:frame_osm_count}.

   \begin{table*}[ht]
    \centering
    \small
    \caption{Statistics of test sets in KITTI and KITTI-360.}
    \label{tab:frame_osm_count}
    \renewcommand{\arraystretch}{1.0}
    \setlength{\tabcolsep}{0.6em}
    \begin{tabular}{c|cc|cccc}
        \toprule
        Dataset & \multicolumn{2}{c|}{KITTI} & \multicolumn{4}{c}{KITTI-360} \\
        \midrule
        Sequence & 00 & 07 & 00 & 05 & 06 & 09 \\
        \midrule
        Point Cloud Frames      & 4541  & 1101  & 10514  & 6291  & 9186  & 13247  \\
        OSM Tiles      & 8782  & 3332   & 12491  &  15080 & 12730  & 13060   \\
        Trajectory Length (m) & 8478 & 3226 & 11612 & 14541 & 12201 & 12570 \\
        \bottomrule
    \end{tabular}
\end{table*}

For the OSM tile sampling strategy, we follow the settings of \citet{cho2022openstreetmap}. During training, OSM tiles are sampled to align with the centers of the corresponding point cloud scans. In contrast, during testing, OSM tiles are uniformly sampled at 1\,\text{m} interval on the ‘highway’ layer in the OSM data.

\section{Implementation Details}\label{sec:Implementation details}

\subsection{Data Pre-processing}\label{appendix: pre-processing}

For both KITTI~\citep{geiger2012we} and KITTI-360~\citep{liao2022kitti} datasets, we use Cylinder3D~\citep{zhu2021cylindrical} pretrained on KITTI to predict 19-class semantic labels (following SemanticKITTI~\citep{behley2019semantickitti}) for each query point cloud.

\begin{figure}[ht]
    \centering
    \includegraphics[width=1.0\linewidth]{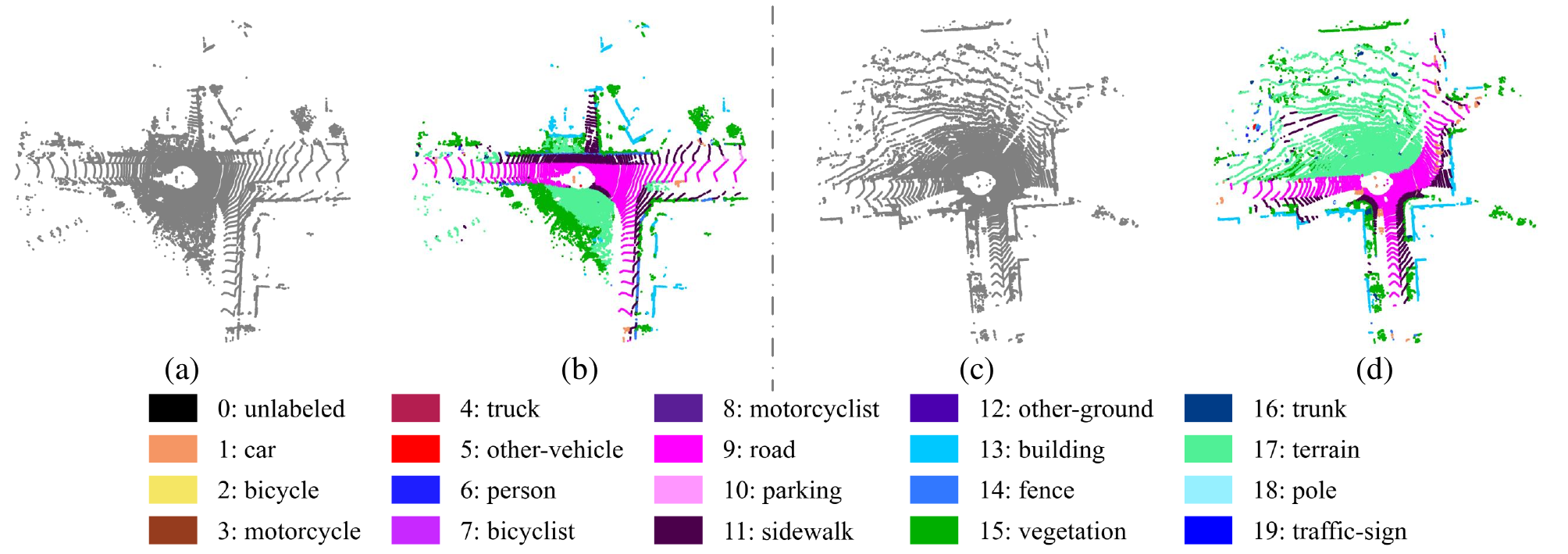}  %
    \caption{Details of semantic point cloud. Figures (a) and (c) display the raw point clouds, while (b) and (d) render them with semantic coloring.}  %
    \label{fig:pc_render}
\end{figure}

\begin{figure}[ht]
    \centering
    \includegraphics[width=0.8\linewidth]{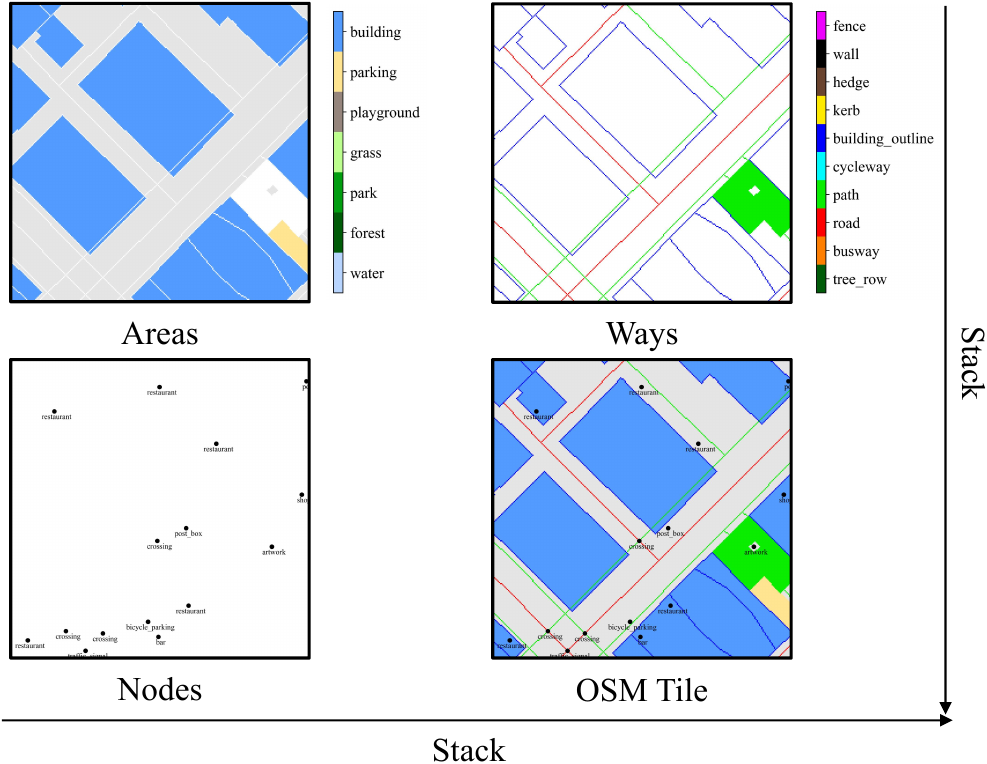}  %
    \caption{Illustration of areas, ways, nodes channels and full OSM tile.}  %
    \label{fig:osm_render}
\end{figure}

 The OSM data is processed through a structured pipeline to generate georeferenced semantic representations. The raw OSM data, comprising various entities, is first categorized into areas, ways and nodes classes according to the hierarchical classification detailed in Tab.~\ref{table:osm}.  Each class is projected onto a local East-North coordinate frame and rasterized into a Cartesian grid with a fixed resolution of  $\Delta_o$ = 50 cm/pixel. As shown in Fig.~\ref{fig:osm_render}, the OSM tiles preserve the semantic and geographic information.

\begin{table}[ht]
\caption{Details of OSM elements.}
\small
\label{table:osm}
\renewcommand{\arraystretch}{1.2}
\begin{tabular}{C{1.5cm}C{11.5cm}}
\hline
Type  & Element                               \\ \hline
Areas & building, parking, playground, grass,
 park, forest, water \\ \hline
Ways  &   Fence, wall, hedge, kerb, building outline, cycleway, path, road, busway, tree row
                                    \\ \hline
Nodes &  parking entrance, street lamp, junction, traffic signal, stop sign, give way sign, bus stop, stop area, crossing, gate, bollard, gas station, bicycle parking, charging station, shop, restaurant, bar, vending machine, pharmacy, tree, stone, ATM, toilets, water fountain, bench, waste basket, post box, artwork, recycling station, clock, fire hydrant, pole, street cabinet
                                     \\ \hline
\end{tabular}
\end{table}

  \subsection{Loss Function}\label{sec:loss function} 
Our OPAL employs the circle loss~\citep{sun2020circle} for optimization. During training, for each query point cloud  $\mathcal{P}$ in a mini-batch, we consider its geographically matching OSM tile as the positive anchor $\mathcal{O}_{pos}$, whereas all other tiles are treated as negative samples $\mathbb{O}_{neg}$. In the shared feature space of the global descriptor, the query point cloud should be close to the positive anchor and far from all the negative anchors. The similarity between a query point cloud descriptor $\boldsymbol{d}_P$ and an OSM tile descriptor $\boldsymbol{d}_O$ is measured through cosine similarity:
\begin{equation}
     s = \frac{<\boldsymbol{d}_P,\boldsymbol{d}_O>}{\|\boldsymbol{d}_P\|\|\boldsymbol{d}_O\|}.
\end{equation}

The optimization objective simultaneously maximizes the similarity $s_{pos}$ between queries and their positive anchors while minimizing similarities $s_{neg}$ with negative anchors. The optimization objective is defined as:
\begin{equation}\label{eq:loss}
\mathcal{L} = \log \left[ 1 + \sum_{i=1}^{|\mathbb{O}_{neg}|} \exp\left(\gamma \alpha_{neg}^i (s_{neg}^i - \Delta_{neg})\right) \cdot \exp\left(-\gamma \alpha_{pos} (s_{pos} - \Delta_{pos})\right) \right]
\end{equation}
where $\alpha_{neg}^i=\mathrm{max}\big(0,s_{neg}^i+\Delta_{neg}\big)$ and $\alpha_{pos}=\mathrm{max}\big(0,1+\Delta_{pos}-s_{pos}\big)$ are dynamic weights for negative and positive mining respectively. The hyperparameters $\Delta_{pos}$ and $\Delta_{neg}$ establish safe margins in the embedding space, while $\gamma$ is a scaling factor controlling gradient sensitivity. %

\subsection{Parameters Setting}\label{appendix:parameter_setting}
Point clouds are filtered to retain points within a range of $3\,\text{m}$ to $50\,\text{m}$, and OSM tiles are of size $H \times W = 100\,\text{m} \times 100\,\text{m}$.  The polar representation consists of $R=480$ rings and $P=360$ sectors. By default, $C_{oem}$ and $C_{pem}$ are set to 16 and 64, respectively, consistent with the configurations adopted in OrienterNet~\citep{sarlin2023orienternet} and PolarNet~\citep{zhang2020polarnet}. In the loss function, positive margin $\Delta_{pos}$, negative margin $\Delta_{neg}$ and scale factor $\gamma$ in the loss function are set to 0.2, 1.8 and 10, respectively.

\section{Additional Results}\label{appendix: results}
\begin{figure*}[ht]
    \centering
    \includegraphics[width=\linewidth]{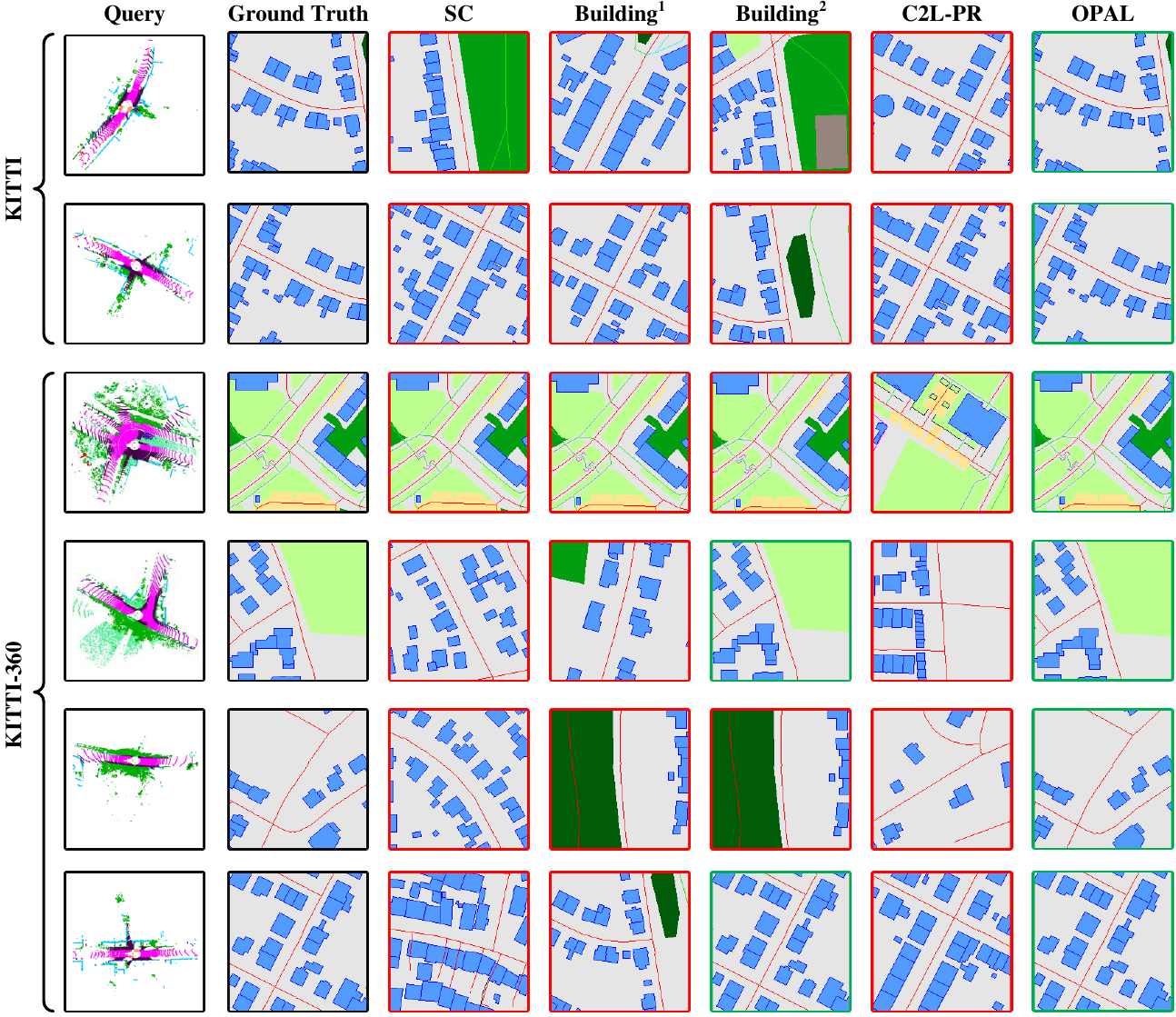}
    \caption{Examples of LiDAR queries and their top-1 retrieved matches on KITTI and KITTI-360 datasets. Red rectangles \( \raisebox{0.5ex}{\textcolor{red}{\fbox{}}} \) represent the wrong retrieved results and green rectangles \( \raisebox{0.5ex}{\textcolor{green}{\fbox{}}} \) represent the correct retrieved results.}  
    \label{fig:Qualitative Append}
\end{figure*}

We provide extensive qualitative results on the KITTI and KITTI-360 datasets, as shown in Fig.~\ref{fig:Qualitative Append}. Compared with both hand-crafted methods~\citep{cho2022openstreetmap,kim2018scan} and learning-based methods~\citep {xu2024c2l},  the proposed OPAL achieves more accurate and robust performance in various scenarios.

\section{Extended Ablation Study}
\textbf{Influence of dynamic objects:} Since OSM data contains only static map elements, whereas query point clouds include numerous dynamic objects (e.g., pedestrians and moving vehicles), this discrepancy may influence localization performance. To examine this effect, we remove dynamic objects from the KITTI dataset using ground-truth semantic labels in the test set and re-evaluate our model. As reported in Tab.~\ref{tab:geo_performance}, OPAL achieves comparable performance with and without dynamic objects, highlighting its robustness to such distractions.

\textbf{Influence of noise in OSM data: } To investigate the effect of noise in OSM data, we introduce random offsets in $[0\text{m},0.5\text{m}]$
 to simulate localization noise, as shown in the last row of Tab.~\ref{tab:geo_performance}. Despite the degraded OSM quality, our method maintains robustness and delivers comparable performance.

\begin{table}[t]
    \centering
    \customsize
    \caption{Ablation studies on the influence of dynamic objects and random noise in OSM data (Top-1 Recall@$K$m retrieved results on the KITTI dataset).}
    \label{tab:geo_performance}
    \renewcommand{\arraystretch}{1.0}
    \setlength{\tabcolsep}{0.37em} 
    \begin{tabular}{@{}c|ccc|ccc@{}}
        \toprule
        \multirow{2}{*}{Setting} & \multicolumn{3}{c|}{KITTI 00} & \multicolumn{3}{c}{KITTI 07} \\
        \cmidrule(lr){2-4} \cmidrule(lr){5-7}
                                 & R@1 & R@5 & R@10 & R@1 & R@5 & R@10 \\
        \midrule
        OPAL                 & 21.82 & 65.78 & 66.40 & 45.41 & 69.85 & 70.30 \\ 
        OPAL w/o Mov. Obj.   & 21.80 & 65.78 & 66.42 & 45.14 & 69.66 & 70.21 \\
        OPAL w/ Random Noise & 21.45 & 65.16 & 65.67 & 43.96 & 68.85 & 68.94 \\
        \bottomrule
    \end{tabular}
\end{table}

\textbf{Descriptor dimension.} Localization accuracy is highly dependent on the dimensionality of the global descriptor. Our method employs 2048-dimensional (2048-D) global descriptors. For handcrafted methods, we follow the official implementations: SC~\citep{kim2018scan} uses a 1200-D descriptor, while Building~\citep{cho2022openstreetmap} adopts 10-D and 360-D descriptors in the first and second stages, respectively. For the learning-based approach, C2L-PR~\citep{xu2024c2l} utilizes 3240-D (point clouds) and 2160-D (OSM) descriptors in stage one, followed by a 288-D descriptor in stage two. As shown in Tab.~\ref{tab:kitti_r5_compact}, we additionally report C2L-PR results with a 2048-D descriptor in stage two for a fair comparison. Under the same descriptor dimensionality, our method achieves a significant performance gain over C2L-PR.

\textbf{Effect of visibility mask.} To provide a comprehensive analysis of the visibility mask, we report the performance of OPAL with and without the mask across all test sequences of the KITTI and KITTI-360 datasets, as shown in the last two rows of Tab.~\ref{tab:kitti_r5_compact}. The results indicate that incorporating the visibility mask consistently improves performance on every sequence, highlighting its effectiveness.

\begin{table}[hbt!]
    \centering
    \customsize
    \caption{Ablation studies on global descriptor dimension (Top-1 Recall@$5$\,m on KITTI (K-) and KITTI-360 (K360-) datasets).}
    \label{tab:kitti_r5_compact}
    \renewcommand{\arraystretch}{1.0}
    \setlength{\tabcolsep}{0.37em} 
    \begin{tabular}{@{}c|c|c|c|c|c|c|c@{}}
        \toprule
        Method & Dim & K-00 & K-07 & K360-00 & K360-05 & K360-06 & K360-09 \\
        \midrule
        C2L-PR       & 288  & 9.69  & 13.17 & 8.23  & 4.72 & 3.35 & 9.59 \\
        C2L-PR       & 2048 & 28.52 & 26.79 & 18.97 & 9.12 & 8.01 & 22.13 \\
        OPAL w/o mask & 2048 & 62.21 & 57.49 & 26.15 & 22.62 & 26.98 & 53.90 \\
        OPAL w/ mask  & 2048 & \textbf{65.78} & \textbf{69.85} & \textbf{42.82} & \textbf{30.49} & \textbf{36.38} & \textbf{60.96} \\
        \bottomrule
    \end{tabular}
\end{table}

\textbf{Computational overhead:} We use the $fvcore$ toolbox to measure the computational overhead of the learning-based baseline, C2L-PR, and our method. C2L-PR requires a total of 535.79\,ms, comprising 0.122\,M parameters, 0.694\,M FLOPs, 1.42\,ms for the learnable second stage, and 534.37\,ms for the handcrafted first stage. In contrast, our method runs in 7.05\,ms, with 88.18\,M parameters and 30.59\,G FLOPs for the whole process.

\end{document}